# PPO-STGNN: A Proximal Policy Optimization Approach with Spatio-Temporal Graph Neural Networks for DAG Task Scheduling in Cloud-Edge-End Computing

Yangshuo Qi[1] , Chenwei Wang[2], Zihan Shen[3] and Songlin Sun[*]

[1] Beijing University of Posts and Telecommunications, Beijing, China
[2] Google, Mountain View, USA
[3] NanJing University, NanJing, China
SLSUN@BUPT.EDU.CN

**Abstract.** With the rapid development of the Internet of Things, computation intensive directed acyclic graph (DAG) tasks have become increasingly common in cloud-edge-end collaborative environments. However, cloud, edge, and end nodes are highly heterogeneous in computing capacity, network bandwidth, and energy consumption, which makes the efficient scheduling of tasks with complex dependencies an NP-hard problem. Traditional heuristic algorithms and conventional reinforcement-learning methods often fail to capture the spatio-temporal dynamics of system resources. This paper proposes PPO-STGNN, a DAG task-scheduling algorithm that integrates proximal policy optimization (PPO) with spatio-temporal graph neural networks (STGNNs). The method uses an STGNN to extract features from both the DAG task topology and the physical cloud-edge-end resource graph, and then optimizes the scheduling policy through PPO to minimize makespan and schedule length ratio (SLR) while improving CPU and memory load balancing. To accelerate convergence, a multi-teacher behavior-cloning mechanism is introduced for pretraining. Experimental results show that PPO-STGNN significantly improves load balancing while maintaining a low completion time, making it suitable for dynamic and heterogeneous cloud-edge-end DAG scheduling scenarios.

## 1 Introduction

With the popularization of IoT and 5G/6G technologies, massive data-processing demands have driven the evolution of computing paradigms toward cloud-edge-end collaborative architectures [1]. In such architectures, computing tasks are offloaded to edge or terminal nodes close to data sources, thereby substantially reducing transmission latency and alleviating backbone-network congestion. Modern intelligent applications, such as video analytics and federated learning, usually consist of multiple subtasks with strict logical and data dependencies. These applications can be mathematically modeled as directed acyclic graphs (DAGs) [2]. However, cloud-edge-end environments exhibit strong spatio-temporal heterogeneity: cloud nodes provide high computational power but introduce cross-layer latency, terminal devices are energy-constrained, and network-link conditions fluctuate dynamically. In this complex heterogeneous environment, efficient DAG task scheduling for minimizing makespan is a typical NP-hard problem [3].

Traditional DAG scheduling algorithms, such as HEFT, depend on static rules and greedy strategies, which limits their adaptability. Deep reinforcement learning (DRL) improves scheduling flexibility [4], but early MLP-based methods lose important DAG topology by flattening task and resource features. Recent GNN-based methods can capture task dependencies [5], [6], yet most of them focus on static topology and ignore temporal resource changes and hierarchical network constraints. Therefore, they remain insufficient for highly dynamic cloud-edge-end workloads [7].

To address these issues, this paper proposes a cloud-edge-end collaborative DAG task-scheduling algorithm based on PPO and STGNN. The main contributions of this work are as follows:

1. A highly heterogeneous spatio-temporal scheduling model is constructed by jointly considering the computing and energy-consumption differences among powerful cloud nodes, medium-capacity edge nodes, and resource-limited end nodes, as well as strict hierarchical topology and link constraints.
2. A PPO-STGNN scheduling architecture is proposed. The STGNN extracts joint spatio-temporal features from the DAG logical graph and the evolution of physical resources, providing the PPO agent with a global perspective for task-node matching.

3. A multi-teacher behavior-cloning pretraining mechanism is introduced. Expert trajectories generated by heuristic algorithms are used for behavior-cloning pretraining, effectively mitigating the cold-start problem of DRL in large-scale joint action spaces.

# 2 Scenario and System Model

## 2.1 Cloud-Edge-End Scenario

The system consists of cloud servers, edge servers, and terminal devices, forming a hierarchical physical topology, as shown in Figure 1.

- End nodes generate DAG tasks. They have weak computing capability and are energy-sensitive, making them suitable for executing lightweight tasks.
- Edge nodes are deployed near terminals at the network edge. Their computing capability is moderate, and they serve as the main execution layer for medium-complexity tasks.
- Cloud nodes have powerful CPU and memory resources and are suitable for processing computation-intensive heavy tasks.

Communication links are configured with different bandwidth and latency constraints. For example, edge-to-cloud links provide relatively high bandwidth but introduce noticeable cross-layer latency; end-to-end links have the lowest bandwidth and the highest delay; and edge nodes within the same domain have better transmission performance.

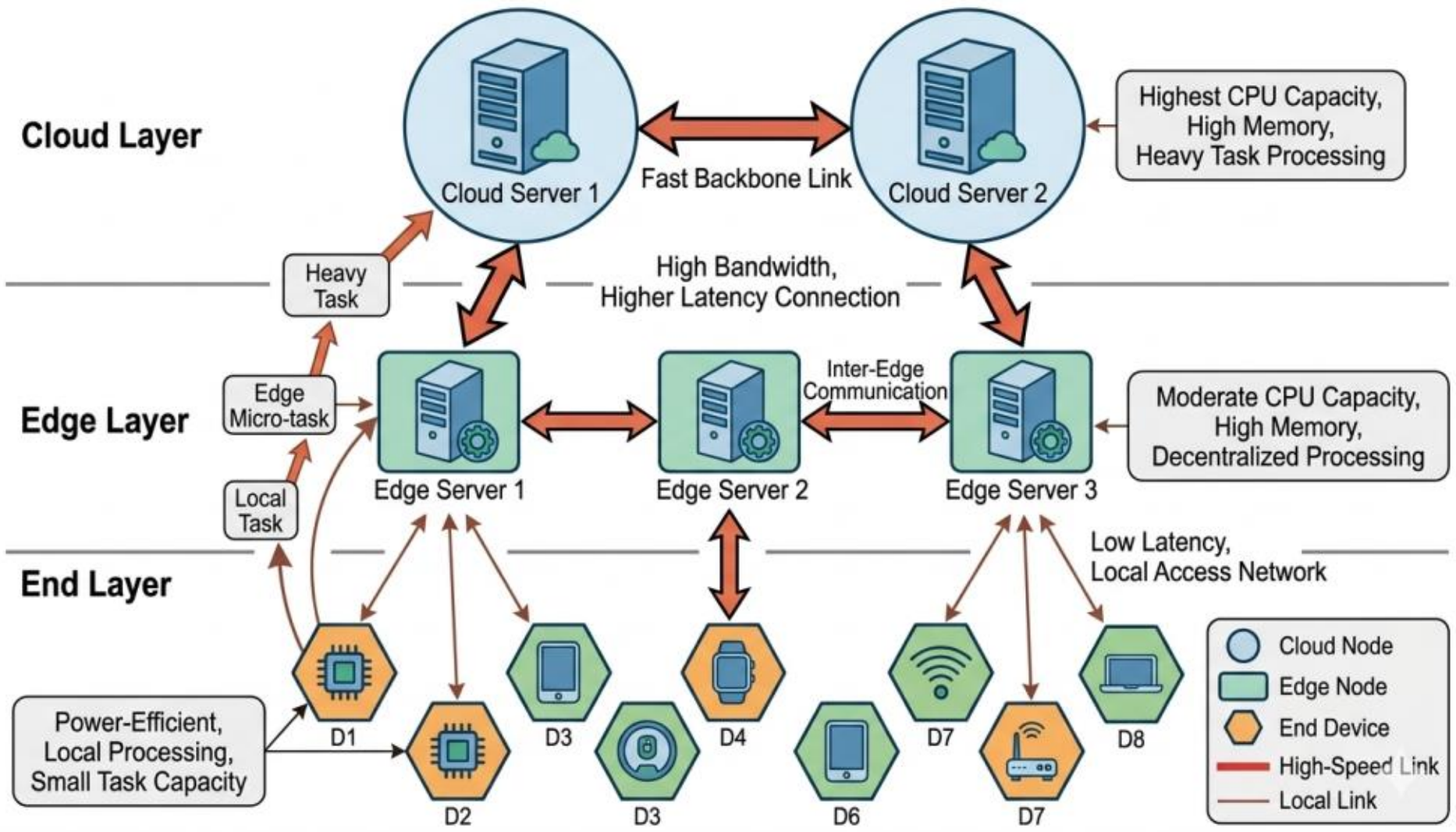

**Fig. 1.** Cloud-edge-end hierarchical computing scenario.

## 2.2 Cloud-Edge-End Network Model

The system is modeled as a hierarchical undirected graph consisting of cloud nodes with strong computing power, edge nodes with moderate computing power, and end nodes that are energy-sensitive:

$$\mathcal{M} = \mathcal{M}_{\text{cloud}} \cup \mathcal{M}_{\text{edge}} \cup \mathcal{M}_{\text{end}}. \tag{1}$$

The computing and memory capacities of node $m_k$ are denoted by $C_k^{\text{cpu}}$ and $C_k^{\text{mem}}$, respectively. Applications are modeled as a set of DAGs $\mathcal{G} = \{G_1, G_2, \dots, G_N\}$, where nodes represent subtasks and edges represent data dependencies. The actual execution time $T_{\text{exec}}(v_i, m_k)$ of task $v_i$ on node $m_k$ is affected by the current effective CPU idle rate $\eta_k(t)$ of the node:

$$T_{\text{exec}}(v_i, m_k) = \frac{w_i}{C_k^{\text{cpu}} \cdot \eta_k(t)}. \tag{2}$$

This paper jointly optimizes the following three core metrics:

1. Makespan ($C_{\max}$): the overall scheduling span of all DAGs.
2. Schedule length ratio (SLR): the ratio between the actual completion time and the execution time of the shortest critical path of the DAG on the optimal node.

3. Load balancing ($L_{\text{CPU}} + L_{\text{Mem}}$): the sum of the variances of normalized CPU and memory loads across nodes.

# 3 Proposed PPO-STGNN Algorithm

DAG task scheduling is a typical Markov decision process (MDP). This paper proposes an end-to-end reinforcement-learning framework that integrates STGNN and PPO, as shown in Figure 2.

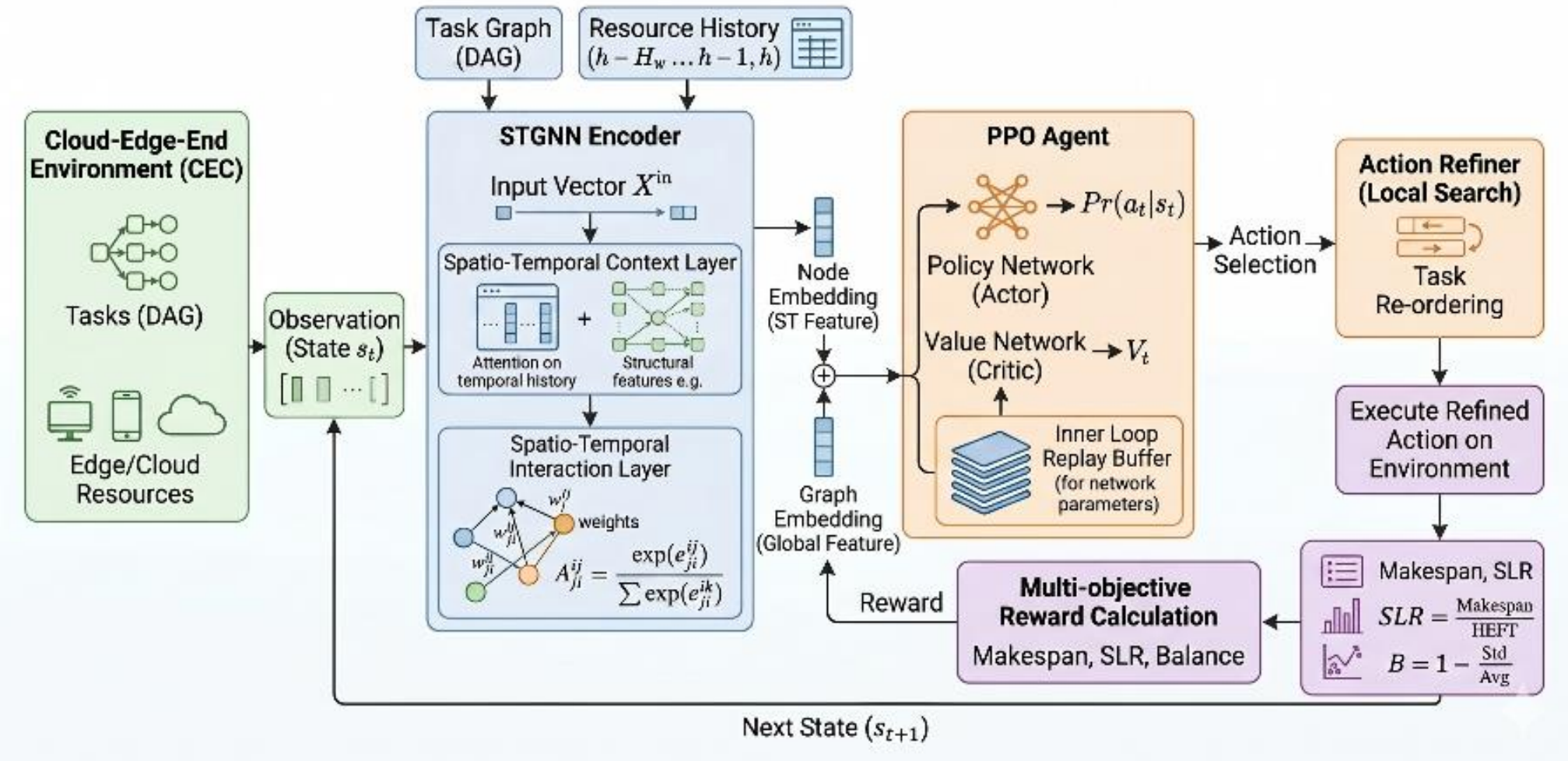

**Fig.2.** Overall workflow of the proposed PPO-STGNN scheduling framework.

## 3.1 State Representation and STGNN Encoder

At each decision step $t$, the system state $s_t$ includes the physical resource state and the logical task-graph state. An STGNN encoder is designed to extract complex topological relationships.

**Physical resource spatio-temporal feature encoding.** Let $X_t^{\text{res}} \in \mathbb{R}^{K\times D_r}$ denote the static and dynamic features of the current $K$ nodes, including CPU/memory idle rate, container-pressure score, and network throughput. First, a graph attention network (GAT) is used to extract spatial connectivity features among nodes:

$$h_{k,\text{spatial}}^{(l+1)} = \sigma\left(\sum_{j\in\mathcal{N}(k)} \alpha_{k,j} W^{(l)} h_{j,\text{spatial}}^{(l)}\right), \tag{3}$$

where the attention weight $\alpha_{k,j}$ is dynamically calculated according to link latency and bandwidth. Then, by combining resource evolution within a historical time window $L$, a gated recurrent unit (GRU) is used to capture temporal dynamics:

$$h_k^{\text{res}} = \text{GRU}\left(h_{k,\text{spatial}}(t), h_{k,\text{spatial}}(t-1), \dots, h_{k,\text{spatial}}(t-L)\right). \tag{4}$$

**DAG task-graph feature encoding.** For the DAG to be scheduled, the depth, in-degree, out-degree, and critical-path score of each task are extracted as the initial node features $X_t^{\text{dag}} \in \mathbb{R}^{N\times D_d}$. A multilayer graph convolutional network (GCN) aggregates predecessor and successor information:

$$H^{\text{dag}} = \text{ReLU}\left(\tilde{D}^{-\frac{1}{2}}\tilde{A}\tilde{D}^{-\frac{1}{2}}X_t^{\text{dag}}W_{\text{dag}}\right), \tag{5}$$

where $\tilde{A}$ is the adjacency matrix with self-loops added. The STGNN finally outputs the joint environmental representation $S_t = \text{Concat}\left(h^{\text{res}}, h^{\text{dag}}\right)$, which is fed into the actor-critic network.

## 3.2 Action Space and Reward Function

Unlike traditional single-step decision making, the proposed algorithm adopts a joint task-node matching action space. The action index is computed as:

$$a_t = \text{Index}(v_i) \times K + \text{Index}(m_k). \tag{6}$$

To satisfy hierarchical-topology constraints, a topological action-mask mechanism is designed to filter illegal routes. The reward function contains a dense step-level penalty that combines response time, transmission cost, and load balancing, denoted by $r_t$, as well as a one-time global terminal reward $R_{\text{term}}$ specifically designed for makespan and SLR.

### 3.3 PPO with Behavior Cloning

To address the slow convergence of DRL, expert trajectories generated by heuristic algorithms such as HEFT are first used for multi-teacher behavior-cloning (BC) supervised pretraining. After pretraining, the agent is fine-tuned online in the real environment using generalized advantage estimation (GAE) and the clipped surrogate objective of PPO to prevent policy collapse.

## 4 Experimental Setup and Result Analysis

### 4.1 Experimental Setup

To verify the DAG workflow-scheduling performance of the proposed PPO-STGNN method in a cloud-edge-end collaborative environment, a heterogeneous computing scenario containing cloud nodes, edge nodes, and end nodes was constructed. Different scheduling methods were compared under the same task set and resource configuration. The experiments focus on three evaluation metrics: task completion time (makespan), schedule length ratio (SLR), and resource load balance.

The experiments consist of two parts. The first part compares PPO-STGNN with traditional baseline scheduling algorithms, including FCFS, LeastLoad, HEFT, and Greedy. The second part further compares the training performance of different deep reinforcement-learning frameworks, including MLP-PPO, PPO-StaticGNN, and PPO-STGNN, to analyze the influence of graph-structure modeling and spatio-temporal feature modeling on scheduling performance.

### 4.2 Comparison with Baseline Methods

As shown in Fig. 3, we compared PPO-STGNN with FCFS, LeastLoad, HEFT, and Greedy. According to the data shown in the Table 1, PPO-STGNN achieves the best performance in both makespan and load balancing.

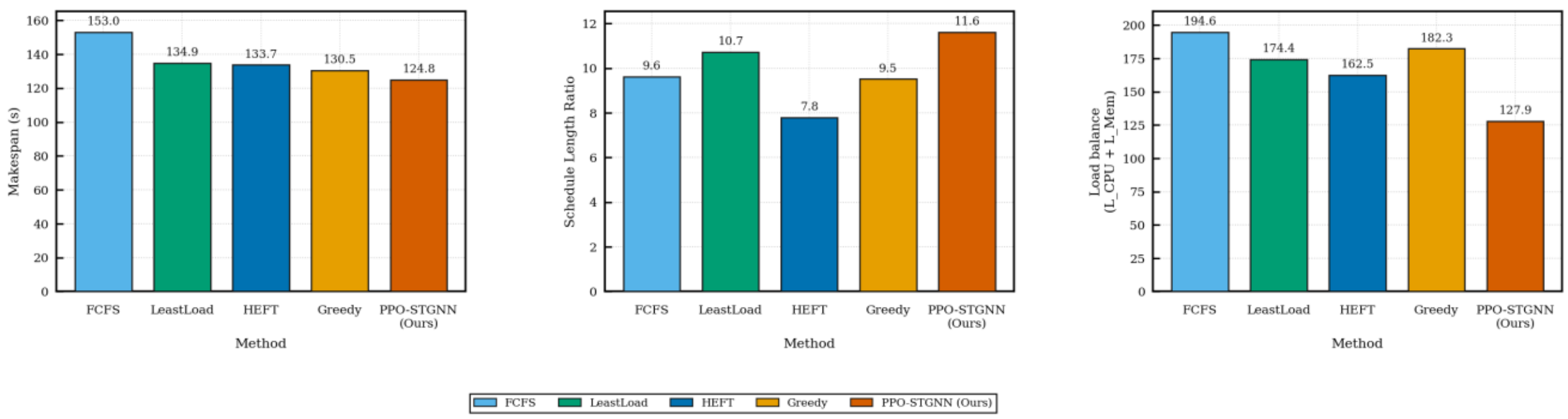

**Figure 3.** Comparison with traditional baseline scheduling algorithms.

In terms of load balancing (LB), PPO-STGNN reduces the LB value to 127.9, representing decreases of approximately 34.3%, 26.7%, and 21.3% compared with FCFS (194.6), LeastLoad (174.4), and the classical DAG scheduling algorithm HEFT (162.5), respectively. This indicates that the proposed algorithm effectively prevents tasks from being excessively concentrated on a few nodes and achieves a more balanced allocation of cloud-edge-end resources.

**Table 1.** Baseline comparison on the same DAG workload and resource configuration.

| Method | Makespan (s) | SLR | LB |
|---|---|---|---|
| FCFS | 153.0 | 9.6 | 194.6 |
| LeastLoad | 134.9 | 10.7 | 174.4 |
| HEFT | 133.7 | 7.8 | 162.5 |

| Method | Makespan (s) | SLR | LB |
|---|---|---|---|
| Greedy | 130.5 | 9.5 | 182.3 |
| PPO-STGNN(Ours) | 124.8 | 11.6 | 127.9 |

In terms of load balancing (LB), PPO-STGNN reduces the LB value to 127.9, representing decreases of approximately 34.3%, 26.7%, and 21.3% compared with FCFS (194.6), LeastLoad (174.4), and the classical DAG scheduling algorithm HEFT (162.5), respectively. This indicates that the proposed algorithm effectively prevents tasks from being excessively concentrated on a few nodes and achieves a more balanced allocation of cloud-edge-end resources.

For makespan, PPO-STGNN obtains the lowest value of 124.8 s, significantly outperforming FCFS (153.0 s) and Greedy (130.5 s). Regarding the schedule length ratio (SLR), HEFT achieves the best result of 7.8 due to its explicit priority and critical-path rules, while PPO-STGNN obtains an SLR of 11.6. Overall, the results show that PPO-STGNN does not focus solely on minimizing a single critical path. Instead, it achieves an optimal trade-off in system resource loading while maintaining a very low global completion time.

### 4.3 Comparison with Different Reinforcement-Learning Frameworks

To further analyze the effect of model structure on scheduling performance, MLP-PPO, PPO-StaticGNN, and PPO-STGNN were compared according to their training trends on the validation set, as shown in Fig. 4.

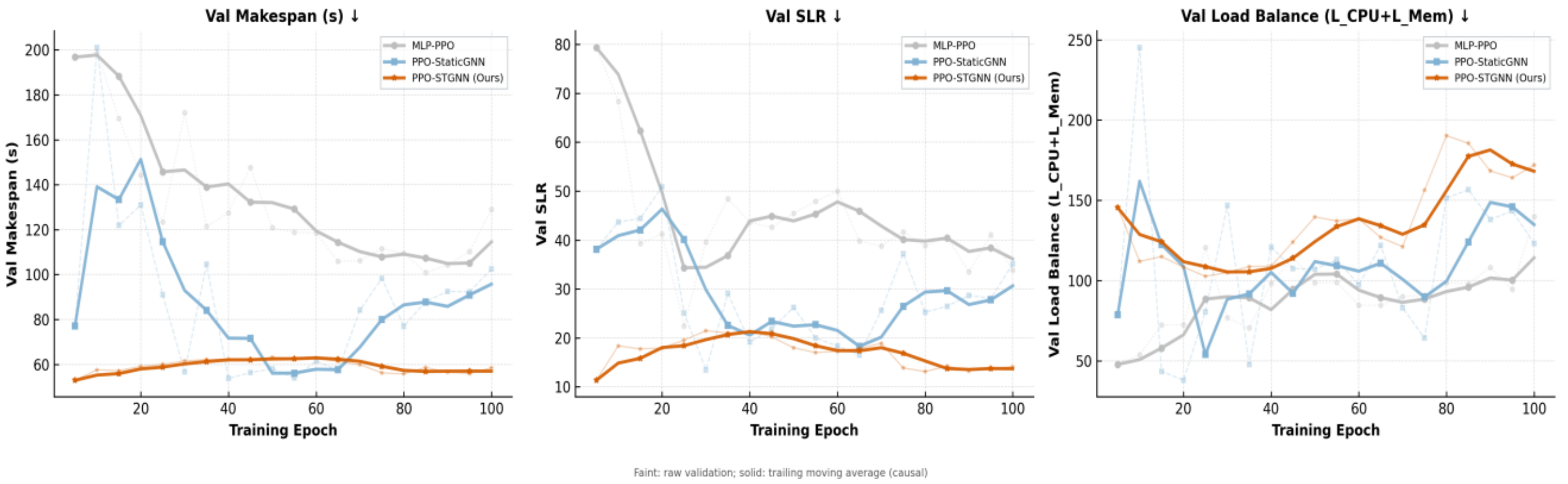

**Fig. 4.** Comparison among MLP-PPO, PPO-StaticGNN, and PPO-STGNN.

In terms of convergence, MLP-PPO, which lacks topology awareness, performs the worst. Its makespan stagnates at approximately 110 s, while its SLR remains around 40. PPO-StaticGNN achieves some improvement by introducing static graph information, but its training process still shows noticeable fluctuations. In contrast, PPO-STGNN rapidly reaches stable convergence by extracting spatio-temporal features, maintaining the lowest makespan below 60 s and an SLR of approximately 15–20.

For multi-objective trade-off, the low load variance of MLP-PPO is actually achieved at the cost of indiscriminately over-dispersing tasks and sacrificing global scheduling efficiency. By comparison, PPO-STGNN keeps the dynamic load within a reasonable range while ensuring clearly superior scheduling efficiency. These results demonstrate that PPO-STGNN successfully overcomes the performance bottlenecks of traditional MLP-based methods and static-graph models in cloud-edge-end environments.

## 5 Conclusion

This paper proposes PPO-STGNN, which models heterogeneous computing resources and workflow tasks as a dynamic resource graph and a DAG, respectively, and captures task-node matching features through spatio-temporal graph representation learning. Experimental results demonstrate that the proposed method effectively improves load balancing while maintaining competitive makespan performance, making it highly suitable for cloud-edge-end collaborative scenarios with dynamically changing resources. Future work will further strengthen the critical-path awareness of the algorithm and explore topology reconstruction caused by terminal-node mobility.